# Digital Twin-Driven Adaptive Sim-to-Real Alignment via Reinforcement Learning for Vibration-Based Bearing Health Monitoring Under Data Scarcity

Jinghan Wang[a], Yanjun Chen[b], Wei Zhang[b], Wentao Wu[a], Tianchen Liu[a], Gaoliang Peng[a]*

[a] Harbin Institute of Technology, China; [b] Eastern Institute of Technology, China

## Abstract:

Vibration-based health monitoring of rotating machinery requires reliable fault diagnosis under operational data constraints, yet condition assessment remains challenged by structural scarcity of fault events and heterogeneous sim-to-real gaps in digital twin-generated signals. Each fault type generates impulses with distinct periodicity, amplitude modulation, and spectral character, making feature-space discrepancies fundamentally heterogeneous across fault classes. Existing domain adaptation methods apply a class-agnostic global transformation that cannot close all fault-specific gaps without distorting inter-class separability, while uniform source-target mixing introduces distributional noise into the data-abundant Normal class. These limitations stem from treating a sequential, state-dependent alignment problem as a one-shot optimization. Each corrective transformation simultaneously reshapes all class distributions, creating state dependencies that static gradient descent cannot resolve. We formulate feature alignment as a continuous-action Markov decision process solved via Proximal Policy Optimization, where the learned policy issues fault-type-specific affine corrections responsive to the current feature-space configuration, with a dual-objective reward balancing gap minimization against separability preservation. An asymmetry-aware strategy reserves real data for the Normal class while augmenting fault classes with policy-aligned simulated samples. Validation across XJTU-SY, CWRU, and a self-built slewing bearing testbed confirms the dominant gain from reinforcement learning-driven alignment, and cross-equipment linear probing achieves 92.8% without encoder retraining, demonstrating transferable monitoring capability.

**Keywords:** Structural health monitoring, bearing fault diagnosis, sim-to-real alignment, digital twin, reinforcement learning, vibration analysis.

## 1. Introduction

Structural health monitoring of rotating machinery has become indispensable for maintaining operational safety across industrial sectors, from wind turbine drivetrains and automated production lines to heavy-duty crane systems. Within these systems, bearings represent both a mechanical cornerstone and a principal failure mode, whose degradation can cascade into catastrophic system-level failures if not detected early through vibration-based condition monitoring [1]. Realizing reliable monitoring under field conditions requires diagnostic algorithms capable of operating with the data constraints that industrial deployment actually imposes, rather than those assumed by laboratory benchmarks. Deep learning has advanced fault diagnosis considerably, yet its deployment in operational monitoring systems remains fundamentally constrained by labeled fault data scarcity. Catastrophic faults are inherently rare, and acquiring representative fault samples under controlled conditions is operationally disruptive and costly. This data scarcity is particularly acute for slewing bearings operating under low-speed heavy-load conditions in excavators, cranes, and turntable systems, where application-specific kinematics and load distributions further limit the generalizability of existing fault signatures.

Digital twin technology offers a compelling solution by leveraging physics-based models that incorporate Hertzian contact mechanics and bearing kinematics to generate large-scale, precisely labeled fault data across diverse operating conditions at negligible cost [2], [3]. This enables development of monitoring algorithms with minimal physical fault data. However, a systematic domain gap exists between simulated and real vibration signals, arising from unmodeled structural dynamics, sensor coupling, and measurement noise.

Two fundamental limitations prevent existing sim-to-real alignment methods from fully bridging this gap. First,

existing methods are designed for static, class-agnostic alignment, including maximum mean discrepancy minimization, correlation alignment, and domain adversarial training, which apply a single global transformation assuming uniform distribution shift across all fault categories [4], [5]. In reality, the sim-to-real gap is structurally heterogeneous. Each fault type produces impulses with distinct periodicity, amplitude modulation, and spectral character, making the feature-space discrepancy for one fault class fundamentally different from that for another. A single global transform cannot simultaneously close all class-specific gaps without distorting inter-class separability.

The second limitation is the symmetric treatment of asymmetric data availability. In real industrial deployments, data from normal operation are abundant while fault data are scarce. Existing methods uniformly mix simulated and real samples across all data classes [6], [7], ignoring this structural data imbalance. Applying the uniform sim-to-real mixing ratio to the Normal class introduces distribution perturbation and degrades the precision of the classification decision boundary.

To address both limitations, this paper proposes a three-stage digital twin framework with reinforcement learning (RL)-driven adaptive feature alignment. In Stage 1, a ResNet-based encoder is pretrained on physics-induced bearing signals to establish a physics-grounded feature manifold. In Stage 2, a proximal policy optimization (PPO) agent formulates sim-to-real feature alignment as a continuous-action Markov decision process (MDP), learning a state-dependent policy that adaptively closes class-specific gaps without collapsing inter-class structure. In Stage 3, the alignment encoder is fine-tuned with an asymmetry-aware strategy that reserves real data exclusively for the Normal class and augments fault classes with RL-aligned simulated samples. The efficacy of the proposed framework is validated on two widely-adopted public benchmarks (XJTU-SY [8], CWRU [9]), as well as a self-built slewing bearing testbed, which represents a mechanically distinct bearing type operating under low-speed heavy-load conditions. The main contributions are as follows.

1) Sim-to-real feature alignment is formulated as a continuous-action Markov decision process solved via Proximal Policy Optimization. The heterogeneous spectral structure across fault classes means each corrective transformation simultaneously reshapes all class distributions, creating sequential state dependencies that one-shot gradient descent cannot resolve, mechanistically justifying the Markovian formulation over static domain adaptation. A dual-objective reward balances domain gap minimization against inter-class separability preservation, preventing boundary collapse during fault-specific alignment.
2) The heterogeneous nature of sim-to-real gaps in bearing fault diagnosis is formally characterized, where each fault type induces structurally distinct feature-space discrepancies arising from differences in impulse periodicity, amplitude modulation, and spectral character. This mechanistic analysis exposes a fundamental incompatibility between the structural heterogeneity of domain gaps and the class-agnostic assumption underlying existing global alignment methods, establishing the theoretical motivation for fault-type-specific adaptive alignment.
3) An asymmetry-aware training strategy is proposed that exploits the structural data availability asymmetry inherent to industrial deployments, reserving real measurements exclusively for the Normal class while augmenting fault classes with policy-aligned simulated samples. A decoupled three-stage architecture with classifier re-initialization accommodates the reshaped feature geometry after RL alignment, enabling cross-equipment monitoring transferability without encoder retraining.

The remainder of this paper is organized as follows. Section II reviews related work. Section III details the proposed framework. Section IV presents experimental results. Section V concludes the paper.

# 2. Related Work

## 2.1 Digital Twin-Based Fault Diagnosis

Early digital twin frameworks for bearing fault diagnosis focused on improving simulation fidelity through refined physical modeling. Multi-body dynamics models incorporating Hertzian contact theory and experimentally validated defect geometries generate vibration signals across fault severities [10]. Subsequent work integrated hybrid physics-data approaches, applying digital twins with contrastive learning to align latent representations [11], while multi-fidelity simulation progressively refines low-cost coarse models with sparse high-fidelity samples [12]. Han et al. [6] recently introduced a multi-scale Kolmogorov-Arnold convolutional network with cross-attention-based feature enhancement, achieving strong cross-condition transfer through adversarial discriminators at multiple semantic scales.

Despite these advances, most existing digital twin-based methods treat the digital twin as a static data generator. Once simulated samples are produced, the source distribution is fixed, and alignment is performed through post-hoc distribution matching. This static paradigm does not exploit the unique advantage of simulation, which is that the source distribution can itself be actively transformed at negligible cost. When the sim-to-real gap is heterogeneous across fault classes, a fixed source distribution cannot simultaneously minimize all class-specific discrepancies without introducing inter-class confusion.

### 2.2 Domain Adaptation for Fault Diagnosis

Domain adaptation (DA) methods learn representations invariant to distribution shift between source and target domains. Early approaches minimized statistical discrepancies. For example Maximum Mean Discrepancy (MMD) penalizes feature distribution distance in reproducing kernel Hilbert spaces [13], while correlation-alignment-based methods match second-order statistics [14]. Recent work has addressed model uncertainty and distribution shift through data-driven adaptive approaches, where online parameter estimation reduces dependencies on complete system models [15]. Adversarial methods, led by domain-adversarial neural network (DANN) [16], introduced domain discriminators trained adversarially to encourage domain-invariant features. Recent work extended adversarial alignment to conditional distributions via conditional domain adversarial network (CDAN) [17] and incorporated prototypical learning [18] to preserve discriminability during alignment.

Chen et al. [19] systematically categorized DA methods into discrepancy-based, adversarial, and reconstruction-based approaches. However, all reviewed methods share a common assumption: the alignment mechanism applies a global transformation derived from aggregate source-target distributional statistics. This assumption is violated in bearing fault diagnosis, where the spectral structure varies systematically across fault types, leading to heterogeneous sim-to-real gaps. A single global transformation that minimizes average domain distance inevitably degrades class separability, as closing the gap for one fault type shifts other classes into misaligned positions.

Furthermore, existing DA methods treat all classes symmetrically during training. In real industrial deployments, the majority of the data are from normal operating conditions, with limited fault data. Applying uniform source-target mixing ratios across all health conditions introduces unnecessary distributional noise into the Normal class, which already possesses sufficient real samples. This reflects a broader limitation in the DA literature: current methods do not account for asymmetric data availability across classes.

### 2.3 Reinforcement Learning in Industrial Condition Monitoring

Reinforcement learning (RL) has been applied to various industrial monitoring tasks, including adaptive sensor scheduling [20], state estimation [21], and predictive maintenance decision-making [22]. Recent work has explored RL for bridging gaps between ideal system models and real-world dynamics through data-informed residual learning, where high-dimensional systems are decomposed into low-dimensional subsystems to improve sample efficiency [23].

In fault diagnosis, RL has been primarily used for discrete optimization: network architecture search formulated layer selection as an MDP [24], while Q-learning enabled feature subset selection [25]. Recent work explored multi-agent RL for hydrostatic transmission control in wind farms [26].

However, RL applications in fault diagnosis have focused on discrete action spaces such as architecture selection or feature subset choice. To the best of our knowledge, RL has not been used for continuous feature-space transformations in domain alignment. This gap is significant: sim-to-real alignment requires continuous affine transformations on high-dimensional features, and the alignment problem exhibits sequential structure where each step alters the feature-space configuration. This naturally fits the MDP formulation, yet existing DA methods perform one-shot optimization without exploiting sequential decision-making.

This work formulates sim-to-real feature alignment as a continuous-action MDP solved via PPO, learning adaptive transformations that close class-specific gaps. We further introduce an asymmetry-aware training strategy that reserves real data for the Normal class while augmenting fault classes with RL-aligned simulated samples.

## 3. Methodology

### 3.1 System Architecture and Framework Overview

The monitoring architecture integrating the testbed, digital twin simulation, RL adaptation module, and diagnostic inference engine is illustrated in Fig. 1.

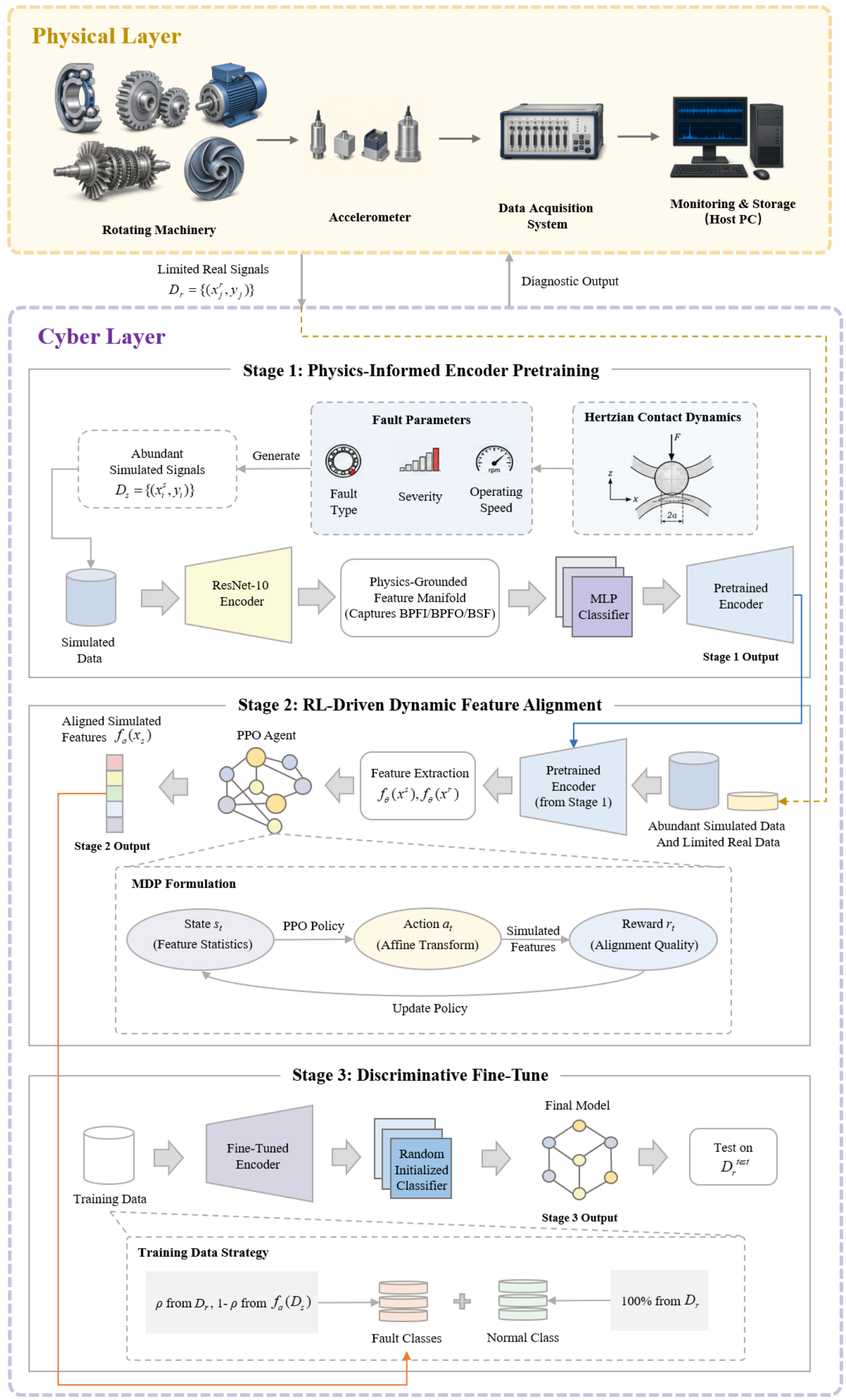

**Fig. 1. Overview of the proposed three-stage framework.**

The physical layer establishes a universal monitoring architecture for rotating machinery fault diagnosis. Various rotating equipment could be chosen as monitored objects, with vibration sensors capturing specific dynamic signatures.

The sensor outputs are processed through a data acquisition system and transmitted to a host PC for storage and preprocessing. This infrastructure generates limited real fault data that serve as input to the cyber layer, while diagnostic outputs from the cyber layer provide condition assessment back to the physical layer. In this work, bearing fault diagnosis serves as a representative case study to demonstrate the framework's effectiveness.

The sim-to-real alignment problem for bearing fault diagnosis can be formally stated as follows. Let $D_s = \{(x_i^s, y_i)\}$ denote the simulated source domain with abundant labeled samples, and $D_r = \{(x_j^r, y_j)\}$ denote the real target domain with limited labeled samples, where the number of simulated samples far exceeds real samples. The input space represents dual-channel vibration signal segments, and the output space comprises four bearing health conditions: Normal, Inner, Outer, and Ball. The objective is to learn a feature encoder $f_\theta$ and a classifier $g_\varphi$ that achieve high classification accuracy on the real test distribution, given that the conditional distributions differ across domains:

$$P_s(x \mid y) \neq P_r(x \mid y), \text{ for all } y \in Y \tag{1}$$

due to unmodeled sensor dynamics and environmental noise.

The cyber layer in Fig. 1 illustrates the proposed three-stage framework. Stage 1 pretrains a ResNet-10 encoder on large-scale simulated data to establish a physics-grounded feature manifold. Stage 2 employs a PPO agent to learn a continuous-action policy that dynamically transforms simulated feature distributions to align with real feature distributions. Unlike static global transformations that apply a fixed mapping, the learned policy adapts to current feature-space configurations, enabling class-specific gap closure. Stage 3 fine-tunes the RL-aligned encoder on an asymmetric dataset, using 100% real data for the Normal class, and a mix of real data with RL-aligned simulated samples for the fault classes.

This staged decomposition exploits the unique advantage of simulation, which is that the source distribution can be actively transformed at negligible cost. By isolating alignment as a separate RL optimization problem, the agent learns a reusable policy that transfers across operating conditions.

## 3.2 Stage 1: Physics-Informed Encoder Pretraining

The first stage establishes a robust feature representation by leveraging physics-based simulation. A bearing dynamics model integrating Hertzian contact mechanics [1] and defect-induced impulse generation generates simulated signals across four health conditions. Each fault type produces characteristic spectral signatures at theoretically predictable frequencies. Fig. 2 presents simulated time-domain waveforms and frequency spectra of the four fault classes. While the simulation captures physically meaningful fault signatures, the amplitude envelope and noise floor systematically deviate from real sensor recordings. This inherent sim-to-real gap motivates the RL-based alignment in Stage 2.

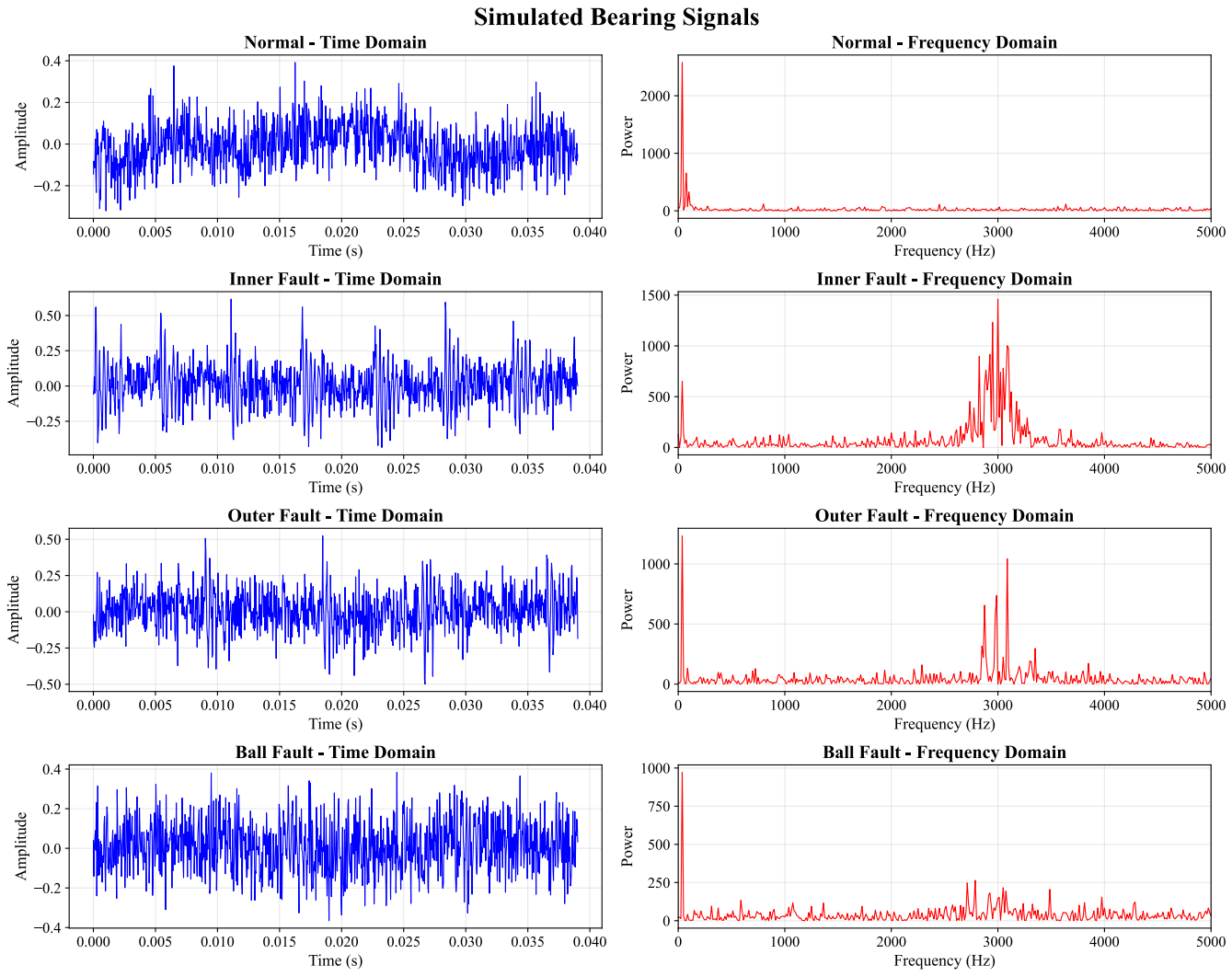

**Fig. 2. Simulated time-domain waveforms and frequency spectra for normal, inner race fault, outer race fault and ball fault states.**

The feature encoder employs a ResNet-10 architecture adapted for one-dimensional dual-channel vibration signals, with input shape (*L*=2048, *C*=2). The architecture comprises four residual blocks with progressively increasing channel dimensions (64→128→256→512), followed by global average pooling and a 128-dimensional feature projection head.

This architecture balances representational capacity with computational efficiency suitable for industrial deployment. The encoder is pretrained on simulated data using cross-entropy loss with label smoothing ($\varepsilon = 0.05$):

$$L_{pretrain} = -\sum_{i=1}^{N_s}\sum_{c=1}^{|Y|} q_i^c \log p_i^c \quad (2)$$

where $N_s$ is the number of simulated training samples, $|Y| = 4$ is the number of fault classes, $q_i^c$ is the smoothed ground-truth label for sample $i$ and class $c$ ($q_i^c = (1-\varepsilon)$ if $y_i=c$, otherwise $q_i^c = \frac{\varepsilon}{|Y|}$), and $p_i^c$ is the predicted probability that sample $i$ belongs to class $c$. This pretraining establishes a feature manifold where fault classes occupy separable regions, providing robust initialization for subsequent alignment.

### 3.3 Stage 2: RL-Driven Dynamic Feature Alignment

This stage constitutes the core technical contribution. The sim-to-real feature alignment problem is formulated as an MDP and solved via PPO.

Table I specifies the MDP five-tuple ($S, A, P, R, \gamma$). The state $s_t$ captures the current feature-space configuration by concatenating batch-level statistics of real and simulated feature distributions: means and standard deviations. The action $a_t$ is a four-dimensional continuous vector representing affine transformation parameters. The transition function deterministically applies the transition:

$$f_a(x^s) = (\Delta\sigma \odot f_\theta(x^s)) + \Delta\mu \, , \quad (3)$$

where $\odot$ denotes element-wise multiplication, $f_\theta(x^s)$ is the feature extracted by the pretrained encoder from simulated sample $x^s$, and $f_a(x^s)$ is the transformed feature.

The 4-dimensional action vector parameterizes physics-motivated corrections. Each of the four action components controls scalar parameters that are broadcast across all 512 feature dimensions to compute the vectorized $\Delta\mu$ and $\Delta\sigma$. The mean shift parameter $\Delta\mu$ controls baseline offset calibration, analogous to direct current (DC) removal in signal conditioning. The variance scaling parameter $\Delta\sigma$ controls amplitude normalization, accounting for damping-induced attenuation. This compact parameterization leverages the insight that sim-to-real discrepancies are complex and non-stationary at the waveform level but can be effectively approximated by affine transformations at the feature level after physics-informed encoding. The broadcasting mechanism applies each of the four scalar action parameters globally across all 512 feature dimensions. This enables the agent to control high-dimensional alignment with a compact and trainable action space.

The reward $r_t$ balances the two objectives:

$$r_t = -MMD(f_a(B_s), f_\theta(B_r)) + \alpha \times Sep(f_a(B_s)) \, , \quad (4)$$

where $B_s$ and $B_r$ denote mini-batches of simulated and real samples respectively, MMD measures multi-kernel maximum mean discrepancy between the transformed simulated feature batch $f_a(B_s)$ and the real feature batch $f_\theta(B_r)$, and Sep measures inter-class margin via mean prototypical distance between class centers. The coefficient $\alpha$ is set to 0.1, prioritizing alignment while maintaining discriminability.

**Table I. IMDP Formulation For Feature Alignment**

| Component | Definition |
|---|---|
| State $s_t$ | $[\mu_r, \sigma_r, \mu_s, \sigma_s] \in \mathbb{R}^{4\times d}$ |
| Action $a_t$ | $(\Delta\mu_1, \Delta\mu_2, \Delta\mu_3, \Delta\mu_4) \in \mathbb{R}^4$ |
| Transition | $f_a(x^s) = (\Delta\sigma \odot f_\theta(x^s)) + \Delta\mu$ |
| Reward $r_t$ | $r_t = -MMD(f_a(B_s), f_\theta(B_r)) + \alpha \times Sep(f_a(B_s))$ |
| Discount $\gamma$ | 0.99 |

The policy is parameterized by an actor-critic architecture. The actor outputs the mean $\mu_a(s)$ and log-standard-deviation log $\sigma_a(s)$ of a Gaussian policy from which actions are sampled during training:

$$\pi_\theta(a \mid s) = N(\mu_a(s), {\sigma_a}^2(s)) \, , \quad (5)$$

where $\pi_\theta$ denotes the policy parameterized by $\theta$, and $N$ represents a Gaussian distribution. The critic estimates the state-value function to reduce variance in policy gradient estimates. PPO updates the policy by maximizing the clipped surrogate objective:

$$L_{PPO} = E_t\left[\min(r_t(\theta)\hat{A}_t,\ clip(r_t(\theta),\ 1-\varepsilon,\ 1+\varepsilon)\hat{A}_t)\right], \tag{6}$$

where $E_t$ denotes expectation over timesteps $t$, $r_t(\theta) = \frac{\mu_\theta(a_t \mid s_t)}{\pi_{\theta old}(a_t \mid s_t)}$ is the probability ratio between new and old policies, $\hat{A}_t$ is the advantage estimate computed via Generalized Advantage Estimation, and $\varepsilon$=0.2 is the clipping threshold that constrains policy updates. PPO is selected for its clipped objective that prevents destructively large policy updates [27], critical for sensitive feature spaces, and its balance between sample complexity and training stability compared to vanilla policy gradient methods.

The continuous action space enables fine-grained, state-dependent adjustments tailored to current feature configurations. This is essential for heterogeneous sim-to-real gaps where different fault classes require distinct alignment magnitudes. The RL agent is trained for 500 episodes, each comprising 200 steps. At each step, mini-batches of 64 samples are drawn from source and target domains, features are extracted via the frozen pretrained encoder, the agent observes $s_t$ and outputs $a_t$, the transformation is applied, and reward $r_t$ is computed. PPO updates occur every 2048 timesteps. The learned policy generates RL-aligned simulated features for Stage 3 training.

### 3.4 Stage 3: Discriminative Fine-Tuning with Asymmetric Data Strategy

The final stage employs an asymmetry-aware training strategy that exploits structural differences in data availability across health conditions. The Normal class uses 100% real data, as normal operation data is abundant in industrial deployments. Fault classes use a configurable real data ratio $\rho$ (set 45% by default), with remaining samples drawn from the RL-aligned simulated distribution. This strategy avoids introducing distributional noise into the Normal class while leveraging simulated data where real fault samples are scarce.

The classifier $g_\varphi$ is randomly initialized rather than inherited from Stage 1. This is necessary because Stage 2's RL alignment fundamentally reshapes the feature space geometry: features that were separated in Stage 1's unaligned space now occupy different relative positions after domain alignment.

The encoder and classifier are jointly fine-tuned using a composite loss function:

$$L_{total} = L_{cls} + \lambda_1(t)L_{proto} + \lambda_2(t)L_{MMD} + \lambda_3(t)L_{DANN}\ , \tag{7}$$

where $L_{cls}$ is the classification cross-entropy loss, $L_{proto}$ is the prototypical loss, $L_{MMD}$ is the multi-kernel maximum mean discrepancy loss, $L_{DANN}$ is the domain adversarial loss, and $\lambda_1(t)$, $\lambda_2(t)$, $\lambda_3(t)$ are time-dependent coefficients that are linearly warmed up from 0 to their maximum values ($\lambda_1^{max}$=0.1, $\lambda_2^{max}$=0.05, $\lambda_3^{max}$=0.01) over the first 20% of training epochs. The prototypical loss $L_{proto}$ pulls same-class features toward learnable class prototypes:

$$L_{proto} = \frac{1}{N}\sum_{i=1}^{N}\left\| f_\theta(x_i) - p_{yi}\right\|^2\ , \tag{8}$$

where $N$ is the batch size, $f_\theta(x_i)$ is the feature extracted from sample $x_i$, and $p_{yi}$ is the learnable prototype vector for the class of sample $i$ (denoted as $y_i$). The prototypes for each class are updated via exponential moving average during training. The domain adversarial loss employs gradient reversal to encourage domain-invariant features.

An asymmetric learning rate strategy preserves the RL-aligned feature space. The encoder learning rate ($1.2\times10^{-5}$) is significantly lower than the classifier learning rate ($2\times10^{-4}$), preventing classifier gradients from overwriting encoder weights.

## 4. Experimental Results

### 4.1 Experimental Setup

Three datasets covering different mechanical configurations are utilized. The XJTU-SY full-life bearing dataset [8] serves as the primary benchmark for the ablation study and real data ratio analysis. XJTU-SY provides run-to-failure vibration records under two operating conditions (35 Hz/12kN and 37.5 Hz/11kN), capturing the complete degradation process from healthy operation through incipient fault development to failure. This full-lifecycle characteristic introduces low-amplitude incipient fault signatures absent from single-fault-state benchmarks, providing a more demanding evaluation. Samples of length 2,048 are extracted by a sliding window with stride 512. The asymmetric training strategy applies only real data for the Normal class and test sets, with a configurable real–simulated mix for fault classes. For cross-dataset robustness evaluation, the CWRU rolling bearing benchmark [9] and a self-built slewing bearing dataset (Handmade) are used under a linear probing protocol, detailed in Section IV-E.

Vibration signals for the three fault states are generated using a Hertzian contact dynamics model. The Normal class uses real measurements throughout all training stages, reflecting the practical reality that normal operating data are readily available.

All experiments use PyTorch 2.0.1 on an NVIDIA RTX 3060 GPU with CUDA 11.8, random seed 42 for reproducibility. Training processes are with AdamW (encoder LR: $1.2\times10^{-5}$, classifier LR: $2\times10^{-4}$), cosine annealing warm restarts, gradient clipping at norm 1.0, and auxiliary loss linear warm-up over 20 epochs.

## 4.2 RL Feature Alignment Analysis

Two complementary visualizations characterize the Stage 2 alignment effect at $\rho = 0.45$. The PCA confidence ellipse plot (Fig.3.) projects real (solid ellipses) and simulated (dashed ellipses) feature distributions onto the first two principal components. Before RL alignment, simulated fault ellipses occupy a completely separate region from their real counterparts, with a fault-averaged inter-domain center distance of 1.109. After alignment, all three fault-class ellipses converge onto their real counterparts, reducing the average center distance to 0.138, an 87.6% reduction. Critically, the three fault classes remain well separated after alignment, confirming that the RL agent closes the domain gap without collapsing class boundaries, a failure mode commonly observed under static MMD alignment.

**PCA Feature Distribution**

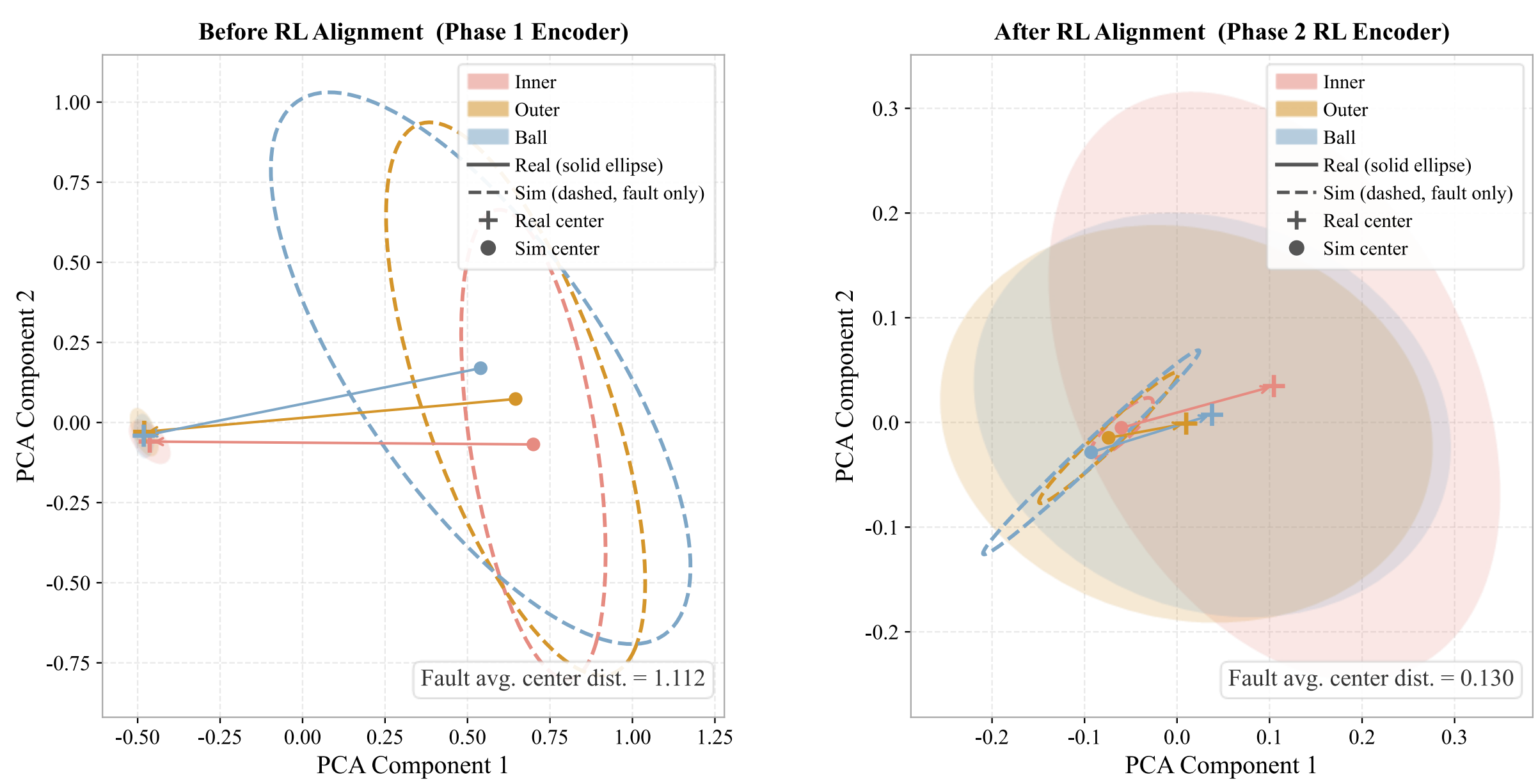

**Fig. 3. CA confidence ellipse plots of real and simulated feature distributions before and after RL alignment.**

The per-class KDE analysis, as shown in Fig. 4, quantifies the distributional shift along the first PCA projection axis. Before alignment, modal positions of simulated and real distributions differ by Δmode = 1.23–1.24 across all fault classes, with Bhattacharyya Coefficients of BC = 0.000–0.038, indicating near-zero distributional overlap. After alignment, the modal distance Δmode drops to 0.006–0.009 and BC rises to 0.482–0.958, confirming near-complete distributional convergence. Taken together, these results demonstrate that the PPO agent learns a fault-type-specific alignment policy that substantially closes the sim-to-real feature gap while preserving class discriminability.

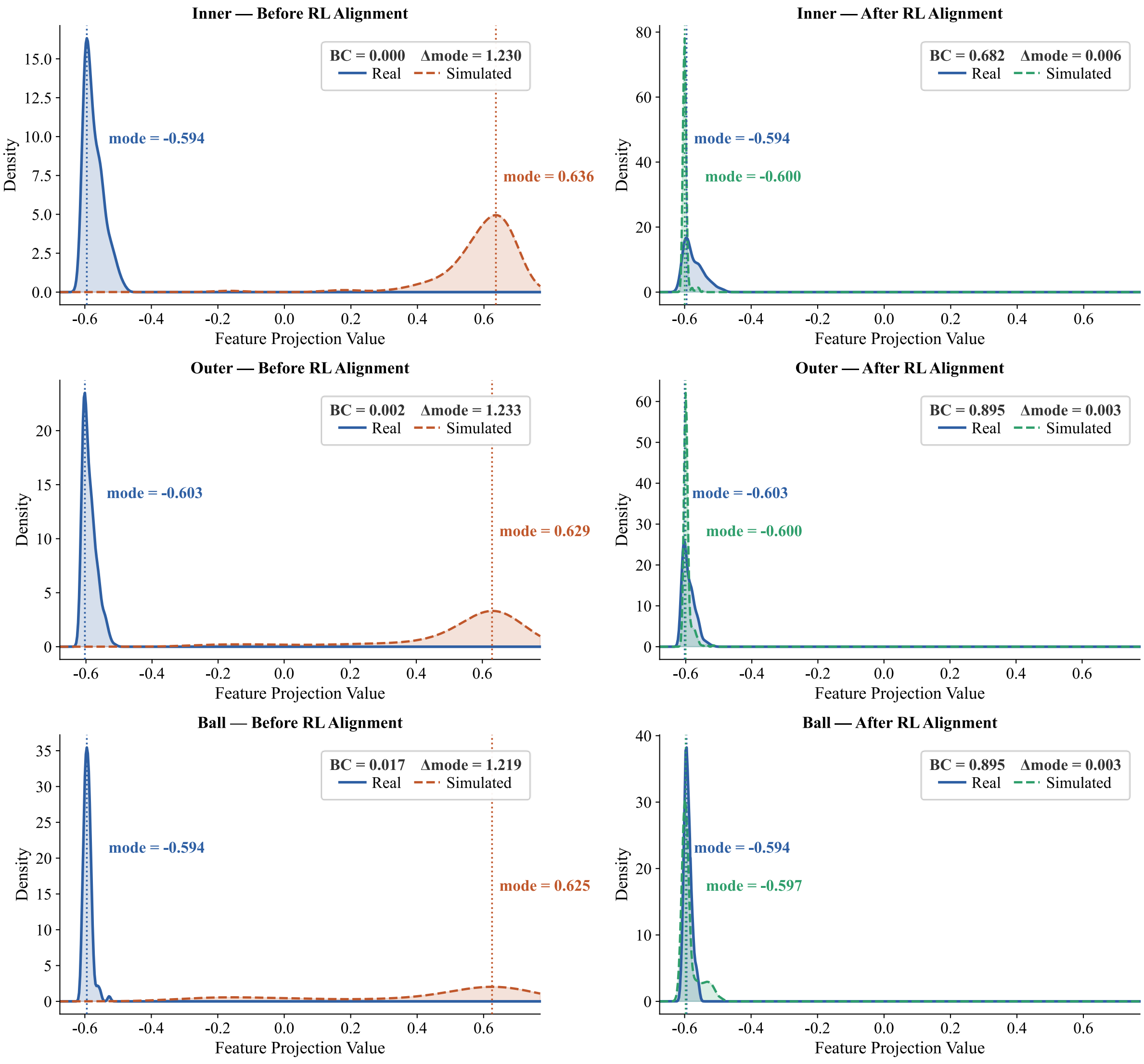

**Fig. 4. Per-class KDE distributions of real and simulated features along the first PCA projection axis before and after RL alignment.**

## 4.3 Ablation Study

Six model variants (M1–M6) are evaluated at $\rho = 0.45$ on the XJTU-SY dataset to quantify the contribution of each framework component. Results are summarized in Table II and illustrated in Figs. 5 and 6.

**Table II. Ablation Study Results on the XJTU-SY Dataset**

| Model | Configuration | Test Acc (%) | Macro-F1 (%) |
|---|---|---|---|
| M1 | Random Enc + Stage 3 | 79.14 | 78.61 |
| M2 | Phase 1 Enc + Stage 3 | 80.28 | 80.68 |
| M3 | Phase 1 + Stage 3+ Static MMD | 81.97 | 82.78 |
| M4 | Phase 2 RL Enc + Stage 3 | 88.13 | 88.03 |
| M5 | Phase 2 RL Enc + Stage 3 + Aux Losses | 89.05 | 88.91 |
| M6 | Full Model (M5 + Class Weights) | 90.47 | 90.47 |

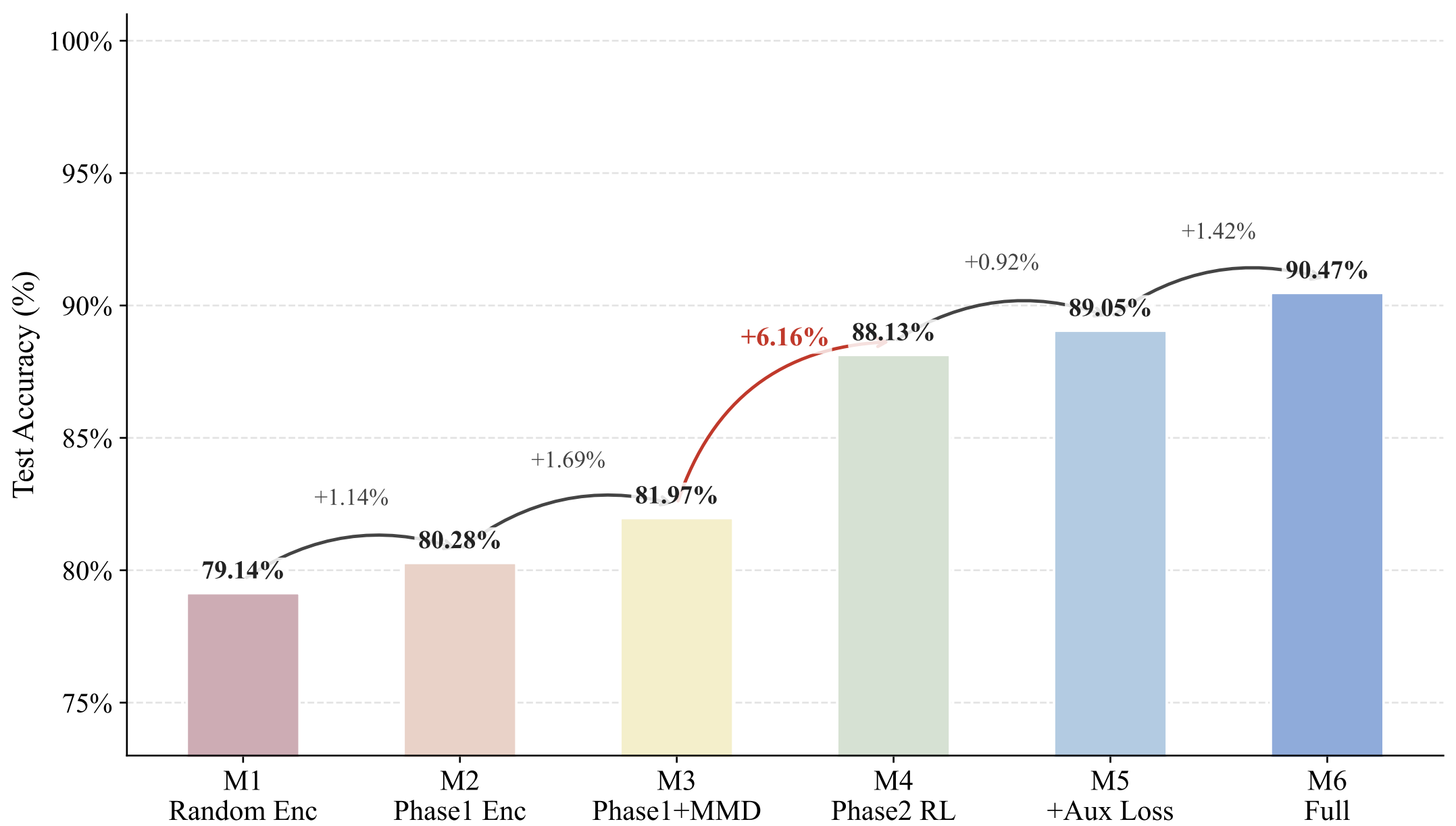

**Fig. 5. Test accuracy of ablation variants M1–M6, with incremental gains annotated between consecutive stages.**

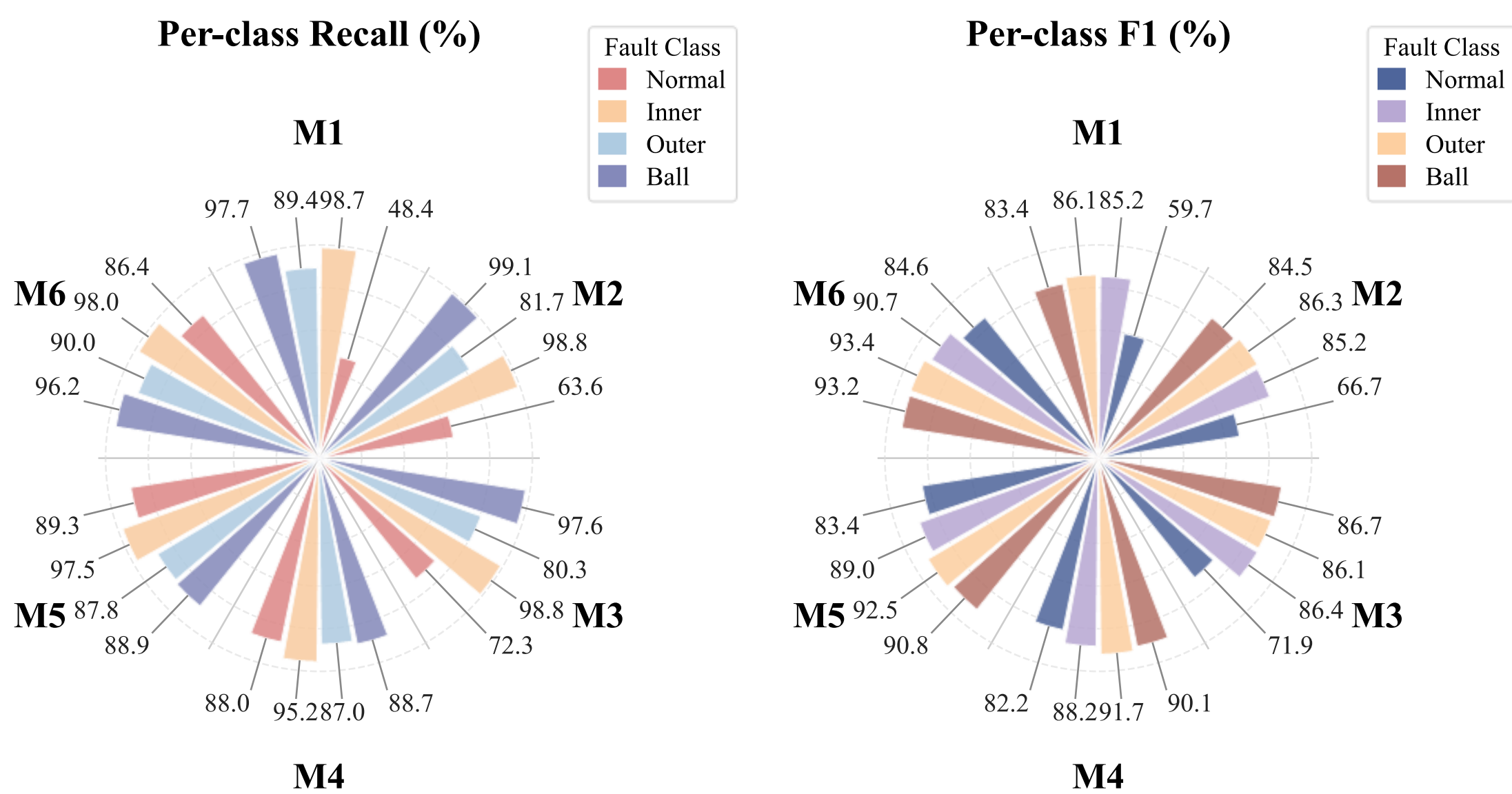

**Fig. 6. Per-class recall and F1 score of ablation variants M1–M6.**

Comparing M1 to M3 reveals the effect of pretraining and static alignment as baselines. Physics-simulated pretraining (M1→M2, +1.14%) and static MMD alignment (M2→M3, +1.69%) yield incremental gains, yet the cumulative improvement is only 2.83%, indicating a fundamental ceiling: a single global transform cannot simultaneously close heterogeneous, fault-type-specific gaps. Normal F1 improves from 59.7% to 71.9% across M1–M3, while fault class scores remain largely stagnant.

Replacing static alignment with the PPO-driven dynamic encoder produces the most decisive improvement (M3→M4, +6.16%), more than twice the total gain across M1–M3. Normal F1 jumps to 82.2% and all fault classes rise above 88%, confirming that the RL agent's adaptive, fault-type-specific policy achieves genuine feature-space convergence.

Auxiliary losses (M4→M5, +0.92%) and asymmetric class-weight fine-tuning (M5→M6, +1.42%) provide consistent refinement, bringing the full model to 90.47% test accuracy and 90.47% macro-F1 with the most balanced per-class performance in the ablation, demonstrating that each stage contributes a distinct and complementary improvement.

## 4.4 Real Data Ratio Analysis

Fig. 7 reports test accuracy on XJTU-SY as the fault-class real ratio $\rho$ varies over {0.1, 0.3, 0.5, 0.7}, with the Normal class always trained on real data only to isolate sensitivity to real fault sample scarcity. The full model achieves 81.7% at $\rho = 0.1$ (10%) and improves monotonically to 90.7% at $\rho = 0.7$ (70%). The Top-5 ensemble and single model curves remain within 1.4% throughout and converge at higher ratios, indicating that the performance gain stems from RL-aligned feature quality rather than model averaging.

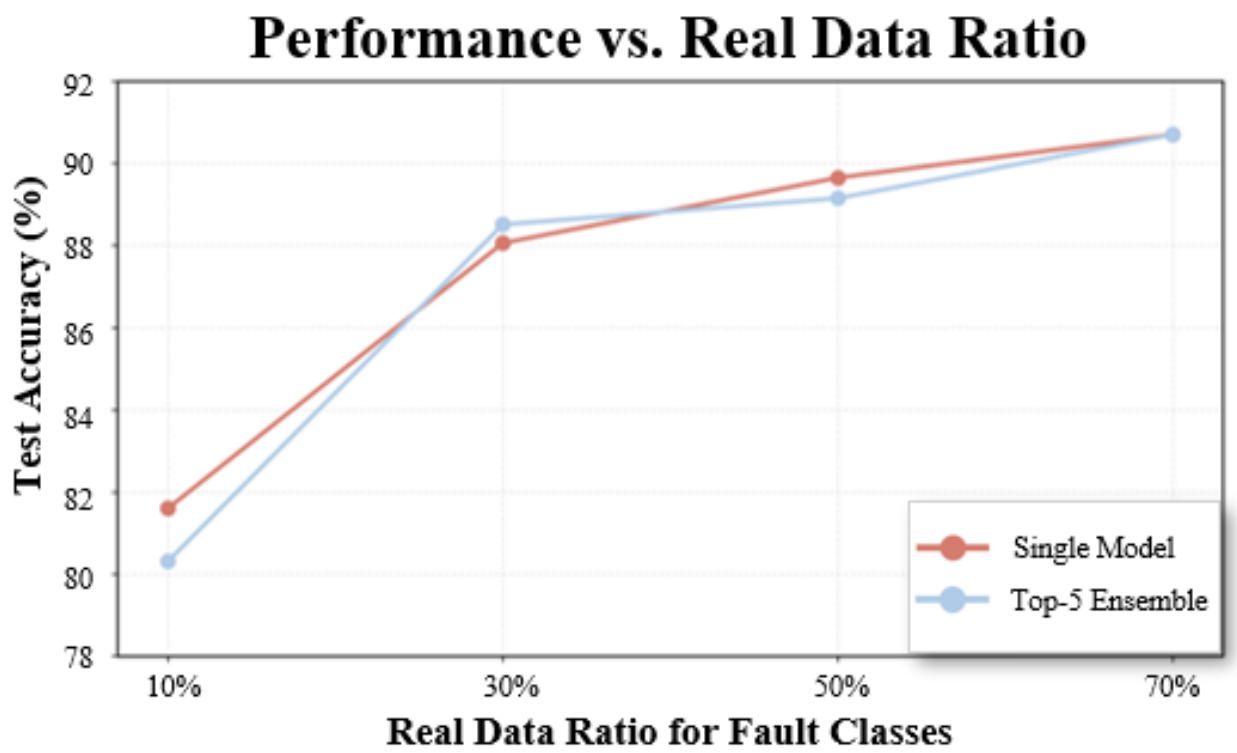

**Fig. 7. Overall test accuracy of the single model and Top-5 ensemble as a function of the fault-class real data ratio $\rho$.**

The per-class trends (Fig. 8) reveal a distinctive pattern at $\rho = 0.1$ (10%): Normal achieves the highest per-class accuracy (92.3%) while Inner Race (72.8%) and Ball fault (69.0%) lag substantially. This does not reflect genuine discriminability but rather the classifier's tendency to over-predict the Normal class: at $\rho = 0.1$, fault class boundaries are insufficiently defined due to scarce real samples, causing a large proportion of fault samples to be misclassified as Normal and inflating its recall while precision drops to 65.3%. As ρ increases, fault class boundaries sharpen and correctly reclaim their feature space, with Inner Race rising from 72.8% to 98.0% at $\rho = 0.5$ (50%) and Normal accuracy stabilizing in the 86–89% range with substantially improved precision, confirming that the framework progressively resolves this class competition effect with increasing RL-aligned real fault data.

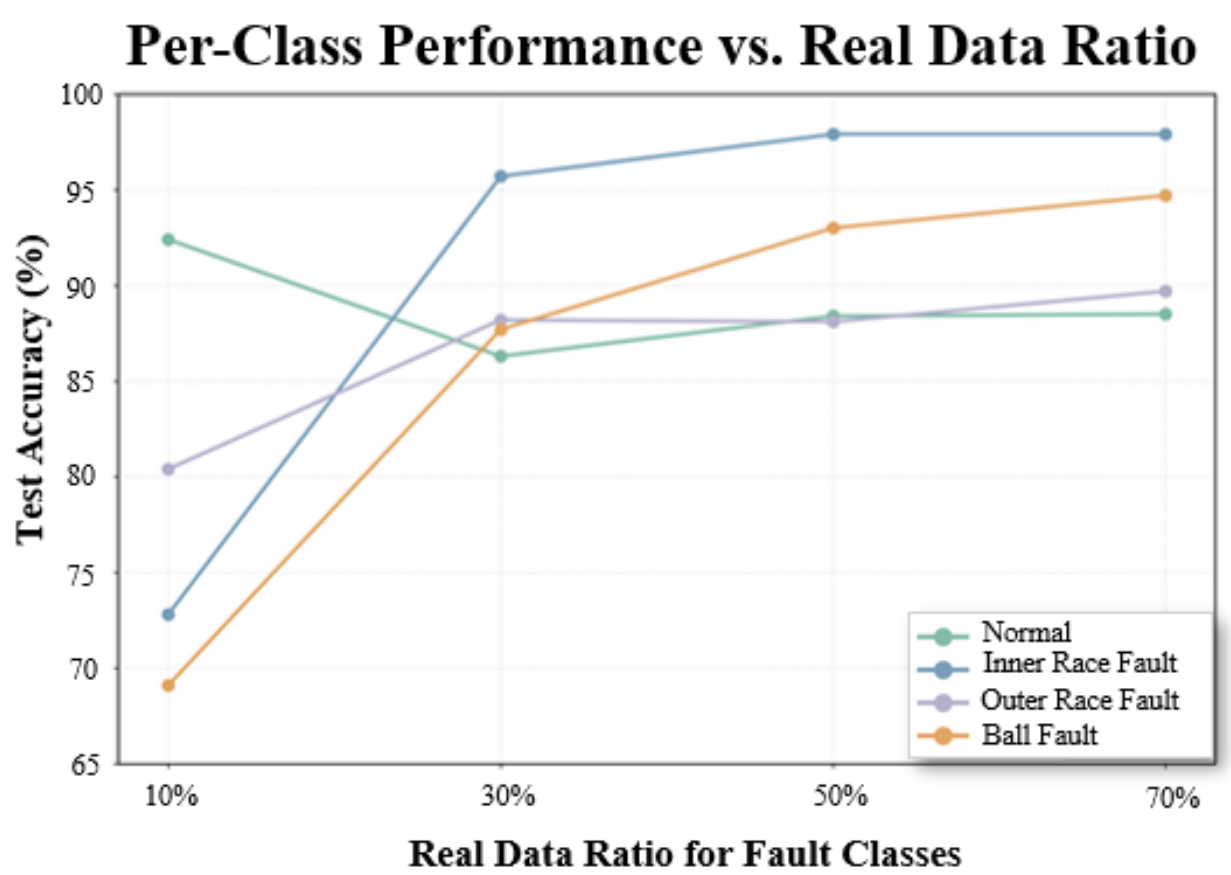

**Fig. 8. Per-class test accuracy as a function of the fault-class real data ratio $\rho$.**

## 4.5 Cross-Dataset Robustness Evaluation

Two cross-dataset evaluations are conducted under a unified linear probing protocol: the XJTU-SY-trained encoder is frozen, and only a lightweight linear classifier is trained and evaluated on each target dataset.

The first evaluation target is a self-built slewing bearing testbed, denoted as "Handmade" in subsequent figures and tables. As shown in Fig. 10(a), the physical testbed employs the system architecture described in Section III-A, deployed at an industrial site to validate the framework under realistic operating conditions. The testbed comprises a 1-HSW320 slewing bearing driven by an LW-100 motor through a PX-8 planetary gearbox at 30:1 reduction ratio, achieving 5 to 20 r/min operational speed under low-speed heavy-load conditions. Two 1A116E ICP accelerometers are orthogonally mounted on the bearing housing with specifications of 10 mV per meter per second squared sensitivity and 2 to 10000 Hz frequency response range. Vibration signals are acquired via an 8-channel data acquisition system sampling at 10 kHz. Data streams to the cyber layer for subsequent processing.

The testbed supports systematic fault injection across four health conditions with three severity levels per fault type, as illustrated in Fig. 10(b). Fault specimens were artificially introduced through controlled machining to simulate progressive degradation stages.

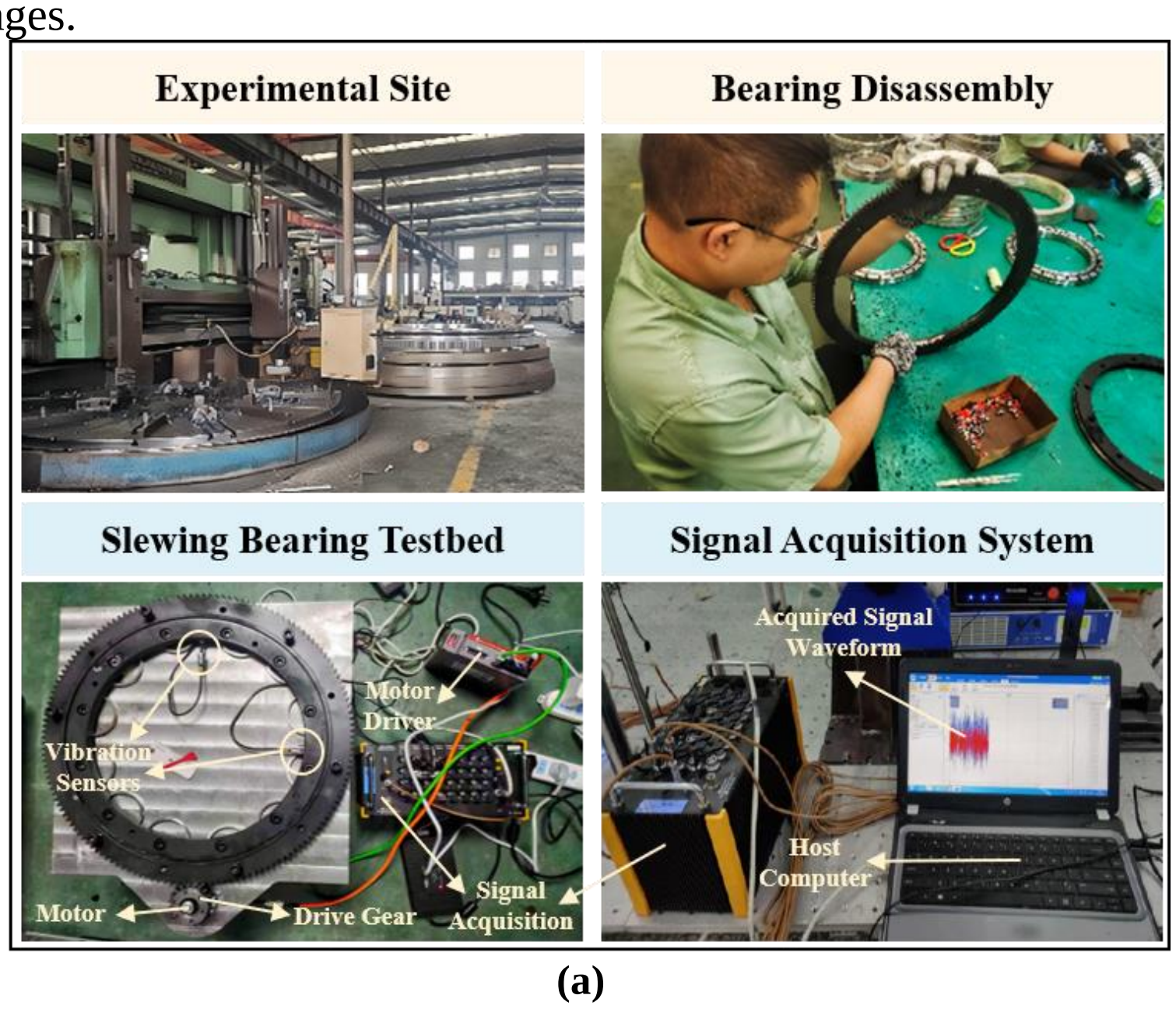

**(a)**

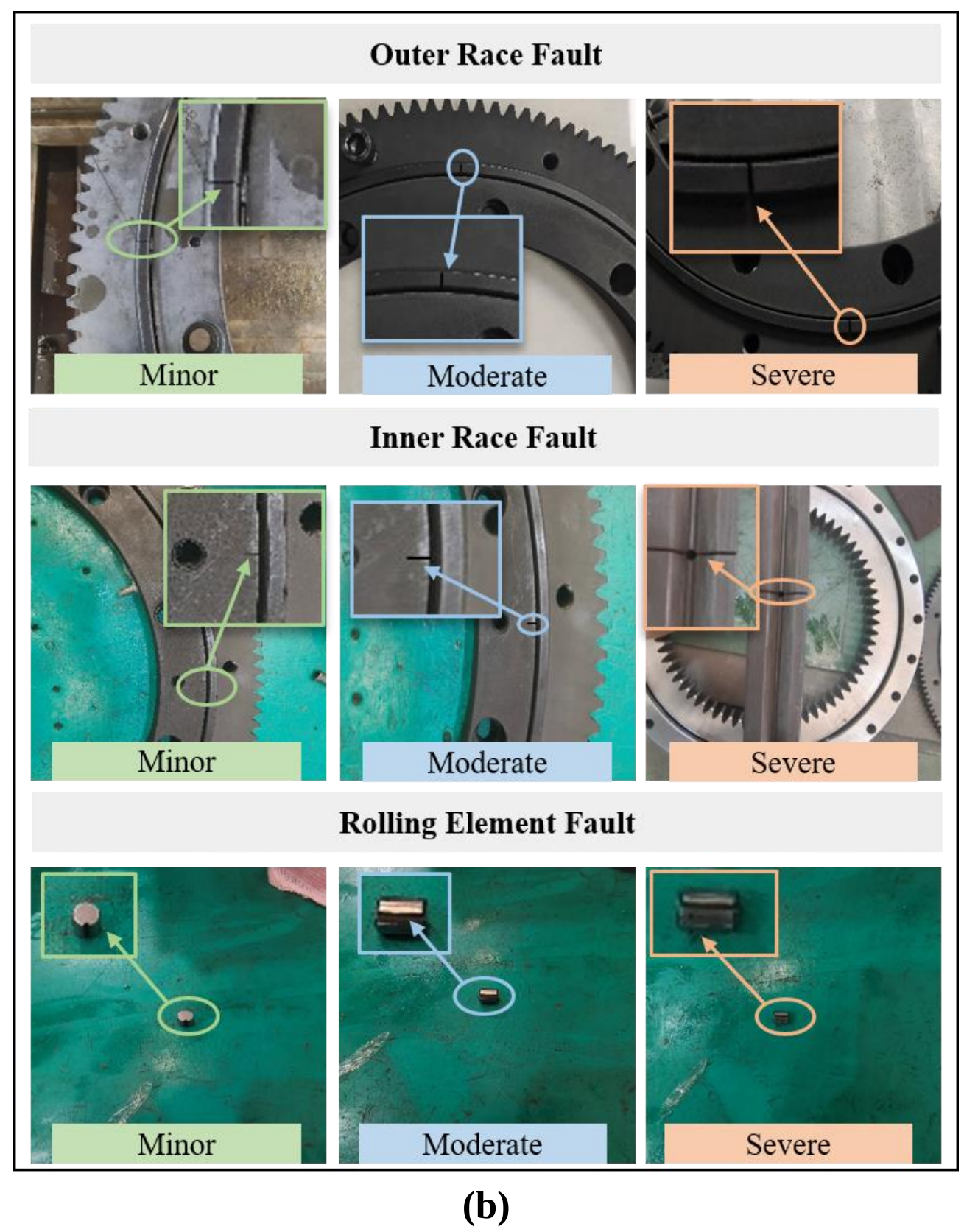

**(b)**

**Fig. 9. Self-built slewing bearing testbed for cross-dataset robustness validation. (a) Experimental platform comprising industrial site deployment (top left), disassembled bearing components (top right), detailed testbed layout (bottom left), and multi-channel data acquisition system with real-time waveform monitoring (bottom right). (b) Representative fault specimens across three fault types and three severity levels, simulating progressive degradation patterns under low-speed heavy-load conditions.**

The second cross-dataset evaluation target is the CWRU rolling bearing benchmark, with its single-channel Drive End signal duplicated to match the dual-channel encoder input. The slewing bearing testbed differs fundamentally from both XJTU-SY and CWRU in roller kinematics, contact mechanics, load distribution, and operating regime (low-speed heavy-load vs. high-speed light-load), providing a stringent out-of-distribution evaluation of the learned representations' transferability across mechanically distinct bearing types.

Quantitative results are summarized in Table III. The full model achieves 92.8% accuracy on CWRU and 79.7% on the Handmade slewing bearing dataset, outperforming all intermediate encoder variants on both targets. Notably, Phase 1 and Phase 2 encoders score below the random baseline on CWRU (86.1% and 85.5% vs. 89.1%), revealing negative transfer from source-domain-specific optimization. The random encoder, unburdened by XJTU-SY distributional biases, learns CWRU-specific patterns more effectively during linear probing. However, Stage 3 fine-tuning resolves this issue: the full model achieves 92.8%, confirming that discriminative fine-tuning removes dataset-specific artifacts while preserving physically meaningful fault signatures. These results demonstrate that the three-stage framework produces representations generalizing across both same-type and cross-mechanical-type targets without any encoder retraining.

**Table III. Cross-Dataset Linear Probing Results on CWRU and Handmade Datasets**

| Encoder | CWRU Acc (%) | CWRU F1 (%) | Handmade Acc (%) | Handmade F1 (%) |
|---|---|---|---|---|
| Random Init | 89.1 | 82.3 | 74.4 | 74.1 |
| Phase 1 (Sim-Pretrained) | 86.1 | 76.8 | 78.0 | 77.7 |
| Phase 2 RL aligned | 85.5 | 75.6 | 77.8 | 77.5 |
| Full Model | 92.8 | 87.8 | 79.7 | 79.5 |

The normalized confusion matrices (Fig. 10) corroborate this, showing that on CWRU, Inner Race reaches 94.9% with minimal confusion, while on the Handmade dataset, Outer Race (68.5%) exhibits the most inter-class confusion. These patterns confirm that the RL-aligned encoder captures equipment-agnostic fault features that transfer across mechanically distinct bearing types, with Outer Race under low-speed heavy-load conditions as the primary remaining challenge.

The per-class radar charts (Fig. 11) show that on CWRU, the full model achieves near-uniform performance across all classes (Inner Race: 94.9%, Normal: 100.0%), while earlier variants show pronounced deficits in Outer Race and Ball. On the Handmade dataset, the performance gap is most pronounced for Inner Race (94.9% on CWRU vs. 77.9% on Handmade, Δ=-17.0%) and Outer Race (83.2% vs. 68.2%, Δ=-15.0%), whose fault signatures under slewing bearing kinematics (larger roller diameter, lower contact angle, distributed load path) differ fundamentally from rolling bearing counterparts. Interestingly, Ball fault accuracy improves on Handmade (79.0% vs. 73.1%), suggesting that the larger roller size amplifies ball defect-induced impulses, making this fault type more distinguishable.

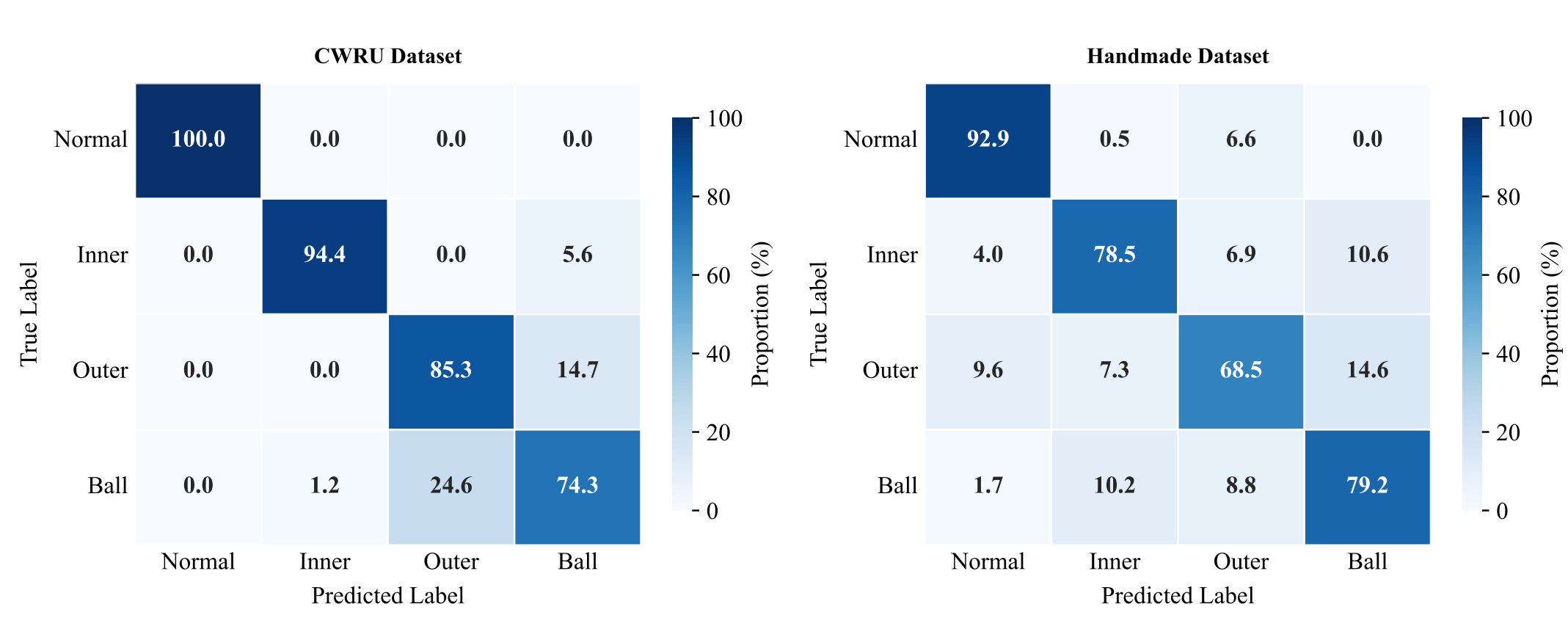

**Fig. 10. Normalized confusion matrices of the full model under linear probing on CWRU and Handmade datasets.**

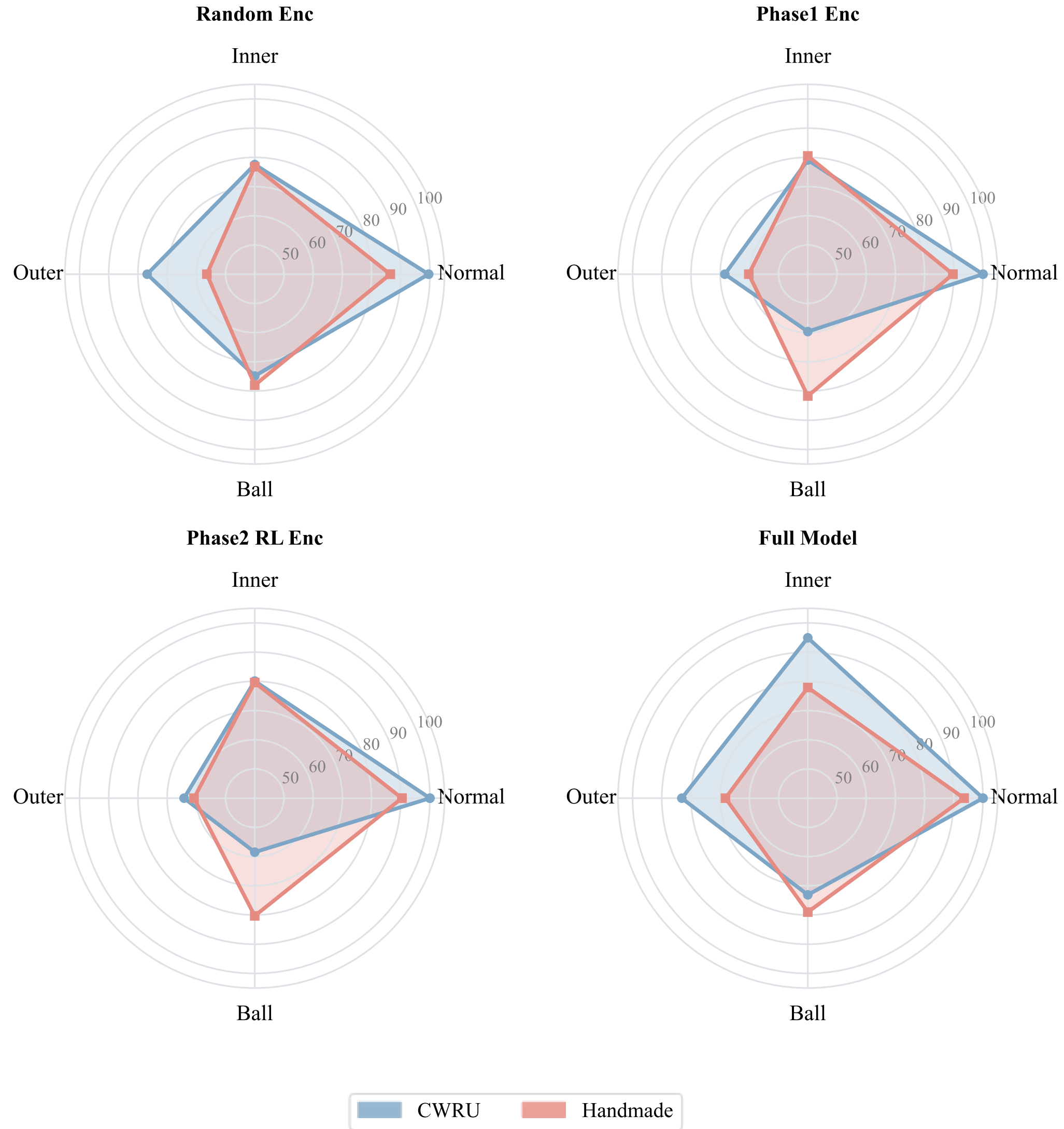

**Fig. 11. Per-class accuracy of four encoder variants under linear probing on CWRU and Handmade datasets.**

# 5. Conclusion

Reliable vibration-based health monitoring of rotating machinery requires diagnostic frameworks effective under real deployment conditions, where fault occurrences are structurally rare and sim-to-real discrepancies are mechanistically heterogeneous across fault classes. Existing global alignment methods are fundamentally incompatible with this heterogeneity, as each fault type induces structurally distinct feature-space discrepancies rooted in differences in impulse periodicity, amplitude modulation, and spectral character, such that a single class-agnostic transformation cannot close all fault-specific gaps without distorting inter-class separability. Recognizing the inherent sequential state dependence of this alignment problem, we formulate sim-to-real feature alignment as a continuous-action Markov decision process solved via Proximal Policy Optimization, enabling fault-type-specific affine transformations adaptive to the current feature-space configuration, which contributes the dominant single performance gain of 6.16 percentage points and achieves 90.47% overall accuracy. An asymmetry-aware strategy and decoupled architecture with classifier re-initialization further address structural data asymmetry and reshaped feature geometry, with cross-equipment linear probing achieving 92.8% accuracy without encoder retraining, confirming transferable monitoring capability of the learned representations.

Currently the proposed framework has two limitations. First, the PPO reward function requires careful design to balance domain gap minimization and class separability preservation. Second, the three-stage pipeline increases training complexity compared to end-to-end methods, though the learned policy is reusable across operating conditions. In future, we will extend the framework to compound fault diagnosis, develop online RL adaptation for non-stationary conditions, and validate generalizability across rotating machinery including gearboxes and electric motors.